# A Comprehensive Study On The Applications of Machine Learning For Diagnosis Of Cancer

Mohnish Chakravarti, Tanay Kothari
{mohnishchakravarti7, tanaykothari1998} @ gmail.com

May 12, 2015

## 1 Abstract

Collectively, lung cancer, breast cancer and melanoma was diagnosed in over 535,340 people out of which, 209,400 deaths were reported [13]. It is estimated that over 600,000 people will be diagnosed with these forms of cancer in 2015. Most of the deaths from lung cancer, breast cancer and melanoma result due to late detection. All of these cancers, if detected early, are 100% curable. In this study, we develop and evaluate algorithms to diagnose Breast cancer, Melanoma, and Lung cancer. In the first part of the study, we employed a normalised Gradient Descent and an Artificial Neural Network to diagnose breast cancer with an overall accuracy of 91% and 95% respectively. In the second part of the study, an artificial neural network coupled with image processing and analysis algorithms was employed to achieve an overall accuracy of 93% A naive mobile based application that allowed people to take diagnostic tests on their phones was developed. Finally, a Support Vector Machine algorithm incorporating image processing and image analysis algorithms was developed to diagnose lung cancer with an accuracy of 94%. All of the aforementioned systems had very low false positive and false negative rates.

## 2 Introduction

Breast cancer, lung cancer and melanoma together account for over 30% of the total cases of cancer reported in 2014. With over 200,000 deaths last year and an expected 300,000 deaths this year, it is imperative that measures are taken to prevent the rise in the deaths occurring collectively due to these diseases.

The 5-year survival rate for lung cancer is 54% However, if not diagnosed early, the chances of survival reduce drastically to 16%. The 5-year survival rate for breast cancer is 86% but it falls to 22% if not diagnosed early. Finally, the 5-year survival rate for melanoma is 95% but it falls to 22% if not diagnosed early. All of these diseases, if diagnosed early, are 100% curable. To diagnose these diseases early, we develop and apply novel machine learning algorithms.

There are several benefits to using Machine Learning algorithms in the field of biomedicine. It eliminates the added dimension of mistakes committed due to human carelessness. The extensive amount of data available online also makes it possible for the algorithms to train themselves and achieve near perfect accuracy - something that is very difficult for any human to achieve. Furthermore, it is easy to replicate and can be shared globally, thus reducing costs associated with logistics and labor force.

For this study, machine learning algorithms such as Support Vector Machine, Artificial Neural Network and Gradient Descent were used. Image processing and analysis was also used to obtain features from images in the case of lung cancer and melanoma.

The next step was feature selection. For breast cancer, features such as the smoothness, concavity, fractal dimension, etc. were used from the WBC database. For lung cancer, image processing and analysis was applied on CT images from the NCBI database to obtain features such as area, convex area, solidity, etc. For melanoma, the ABCD standard prescribed by dermatologists was used. The images were obtained through the $PH^2$ dataset.

**Outline** The remainder of this article is organised as follows. Section 3 contains our research, methods and findings for detecting lung cancer. Section 4 contains our research, methods and findings for detecting melanoma. Section 5 contains our research, methods and findings for detecting breast cancer. Section 6 contains our conclusions. Section 7 and Section 8 present the acknowledgements and the references respectively.

# 3 Lung Cancer

Support Vector Machine along with image processing and image analysis algorithms were used in this part of the study.

## 3.1 Image pre-processing, processing and segmentation

Total variation denoising, which is effective at preserving edges while smoothing noise in flat regions, was used in the pre-processing stage

The optimal thresholding proposed in [17] is applied to the pre-processed image to obtain a segmentation threshold.

$$T_{i+1} = \frac{1}{2}(u_a + u_b)$$

In this equation, $T_i$ is the segment threshold after the $i^{th}$ step. To choose a new segmentation threshold, we apply $T_i$ to the image to separate the pixels into body and nobody pixels. Let $u_a$ and $u_b$ be the mean grey level of body and nobody pixels.

Pixels with density lower than the threshold value are assigned the value 1 and the other pixels are assigned the value 0. The remaining non-body pixels are eliminated through morphological closing.

### 3.2 Feature Selection

After the ROI (Region Of Interest) was obtained, GLCM (Gray Level Co-occurrence Matrix) was used to extract the features. The features extracted and the methods used are shown in the table. (Table 1) [20]

| Feature | Formula | Function |
|---|---|---|
| Area (A) | - | Total number of pixels in the ROI |
| Convex Area (CA) | - | Total number of pixels in the convex region of the ROI |
| Equivalent Diameter (ED) | $ED = \frac{\sqrt{4}+A}{A}$ | Area of circle equal to the area of the ROI |
| Solidity (S) | $S = \frac{A}{CA}$ | Ratio of Area to Convex Area |
| Energy (E) | $E = \sum_{0}^{n} p^2(i,j)$ | Summation of the squared elements in the GLCM |
| Contast (C) | $C = \sum_{i}^{N} \sum_{j}^{N} (i-j)^2 p(i,j)$ | Measure of contrast between intensity of adjacent pixels over the whole ROI |
| Eccentricity (EC) | - | Ratio of distance between the foci of the ellipse and its major axis length |
| Homogeneity (H) | $H = \sum_{i}^{j} \frac{p(i,j)}{1+\lvert i+j \rvert}$ | Measure of closeness of the distribution of elements in the GLCM to the GLCM of each ROI |
| Correlation (CO) | $CO = \sum_{i}^{N} \sum_{j}^{N} \frac{p(i,j) - \mu r.\mu c}{\sigma r.\sigma c}$ | Measure of correlation of pixel to its neighbor over the ROI |

### 3.3 Support Vector Machine Development

A linear classifier was used. We chose the hyperplane such that the distance from it to the nearest data point on each side is maximized. The linear classifier that such a hyperplane defines is known as a maximum classifier.

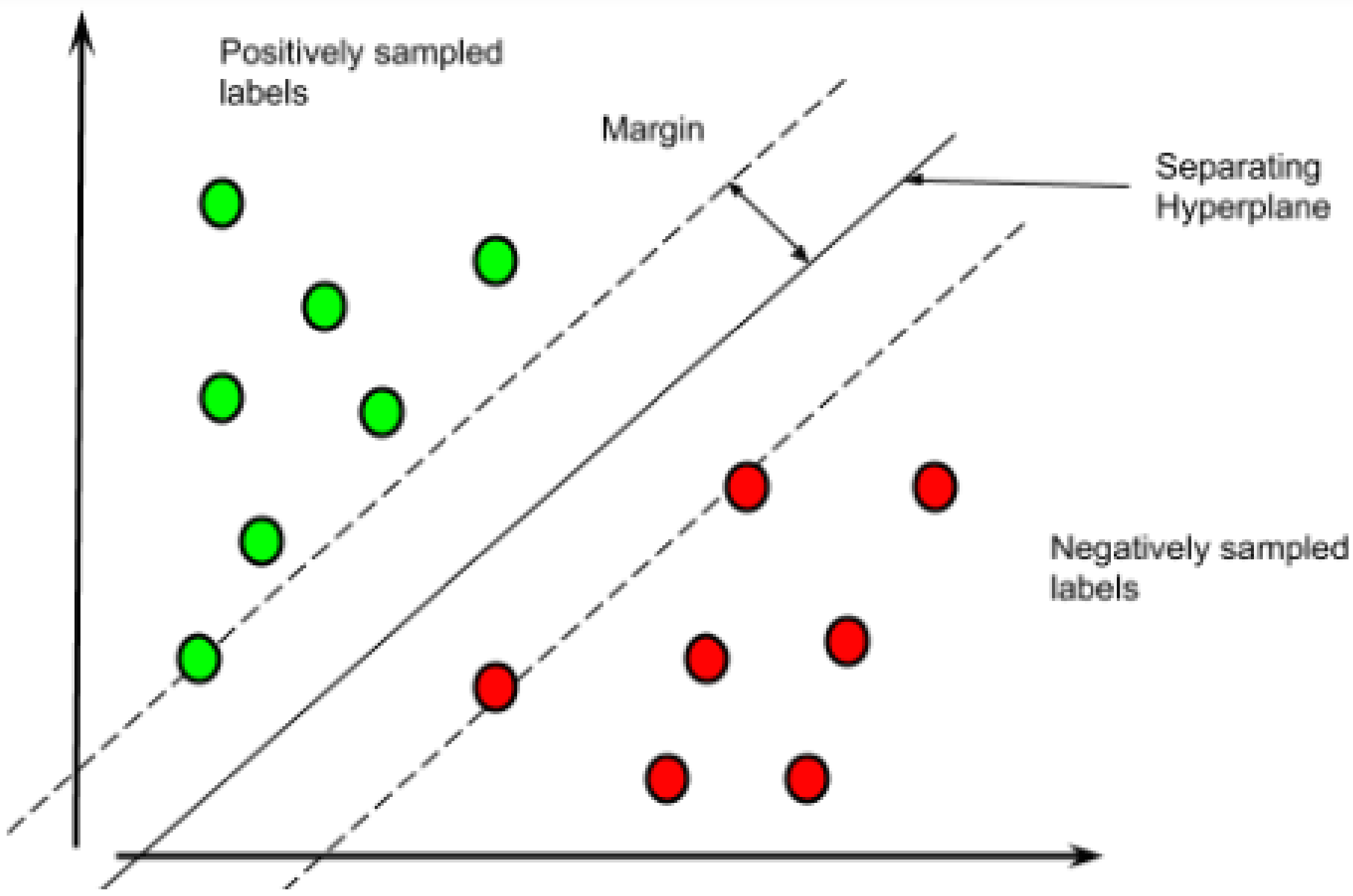

Figure 1: An example of a maximum margin classifier

The CT images used were obtained from NCBI's online database.

### 3.4 Results

Over 197 images were tested on the SVM. The classifier only chose 2 out of the 9 features available for classification at any given time. The most reliable features were Area and Convex Area.

A breakup of 70 : 30 was used for training and testing of our algorithm. The data presented here is from the testing period of the algorithm.

Based on the classification, the tumour was either diagnosed as benign or malignant. The SVM delivered an accuracy of over 94%, mis-diagnosing only 11 of the images. 5 were false positive and 6 were false negative.

| **Output** | **Significance** |
|---|---|
| 0 | Benign |
| 1 | Malignant |

The overall accuracy delivered by the SVM was, as noted earlier, 94%

The results have been summarised below (Figure 2).

<table>
<tr><td rowspan="6">Overall Outcome</td><td colspan="5">Overall Results</td></tr>
<tr><td colspan="2">Positive</td><td colspan="2">Negative</td><td rowspan="2">Sensitivity 95.56%</td></tr>
<tr><td>TP : 129</td><td>FP : 5</td><td>TN : 59</td><td>FN : 6</td></tr>
<tr><td colspan="4" rowspan="2">Inconclusive : 0</td><td>Specificity 92.12%</td></tr>
<tr><td rowspan="2">Matthews Correlation Coefficient<br>0.874</td></tr>
<tr><td colspan="2">Positive Prediction Rate<br>96.23%</td><td colspan="2">Negative Prediction Rate<br>90.77%</td></tr>
</table>

Figure 2: Summary of the results from the Lung Cancer Classifier

# 4 Melanoma

Artificial Neural Networks along with image processing and image analysis algorithms were used in this part of the study.

## 4.1 Image pre-processing, processing and segmentation

Noise was removed using morphological closing. An accurate border was determined. The gradient function was used.

## 4.2 Feature Selection

Features such as asymmetry along the minor axes, asymmetry along the major axis, border irregularity, entropy, color variation, diameter, etc. were extracted using built-in or custom coded functions. The asymmetry of the lesion was determined by overlapping the halves of the images on their major and minor axes. To accurately determine the border in-built functions were used.

## 4.3 Artificial Neural Network Development

The neural network was developed using an in-built toolbox in MATLAB. The features extracted were used as input in the ANN.

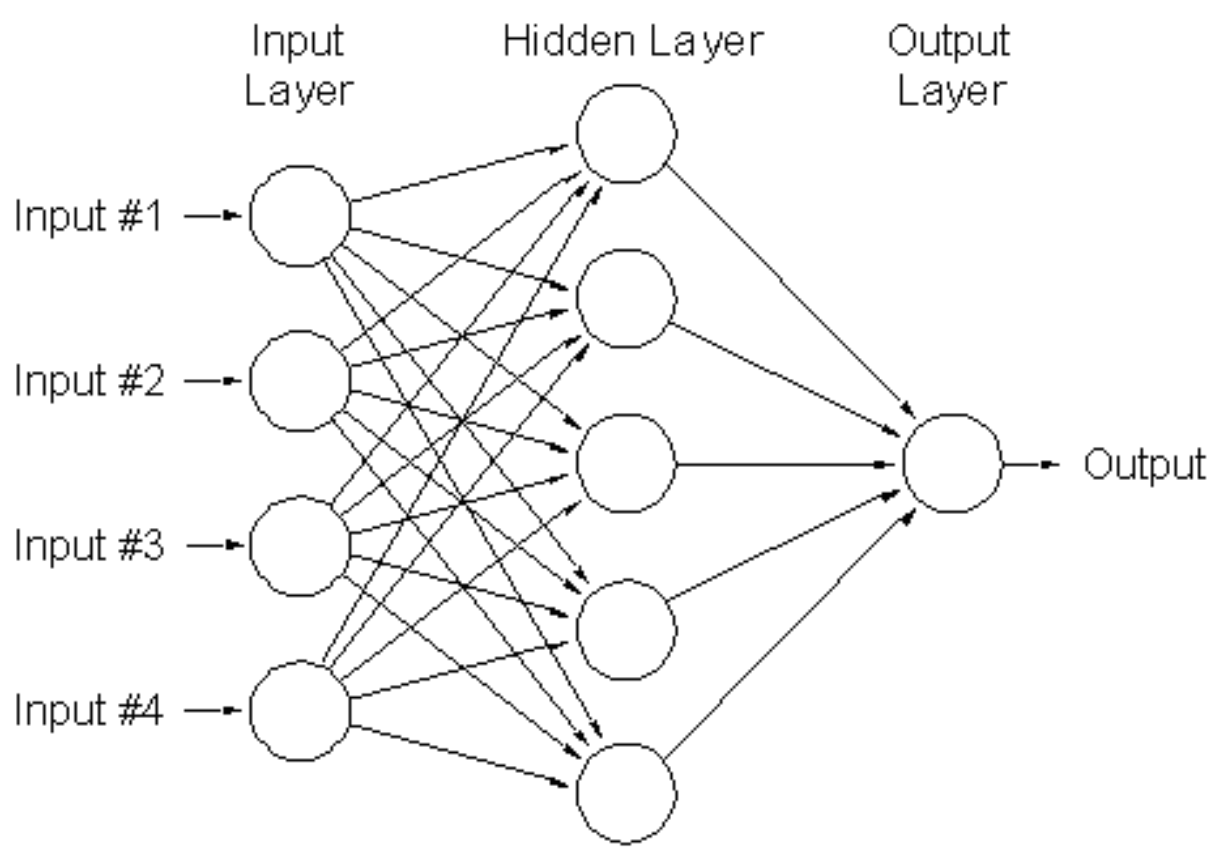

Figure 3: An example of the structure of an Artificial Neural Network with multiple layers.

## 4.4 Mobile Application

The ANN was converted into a standalone service and deployed on the web. It was then integrated into the mobile application.

The Melanomore application is currently under development and further details will be released shortly.

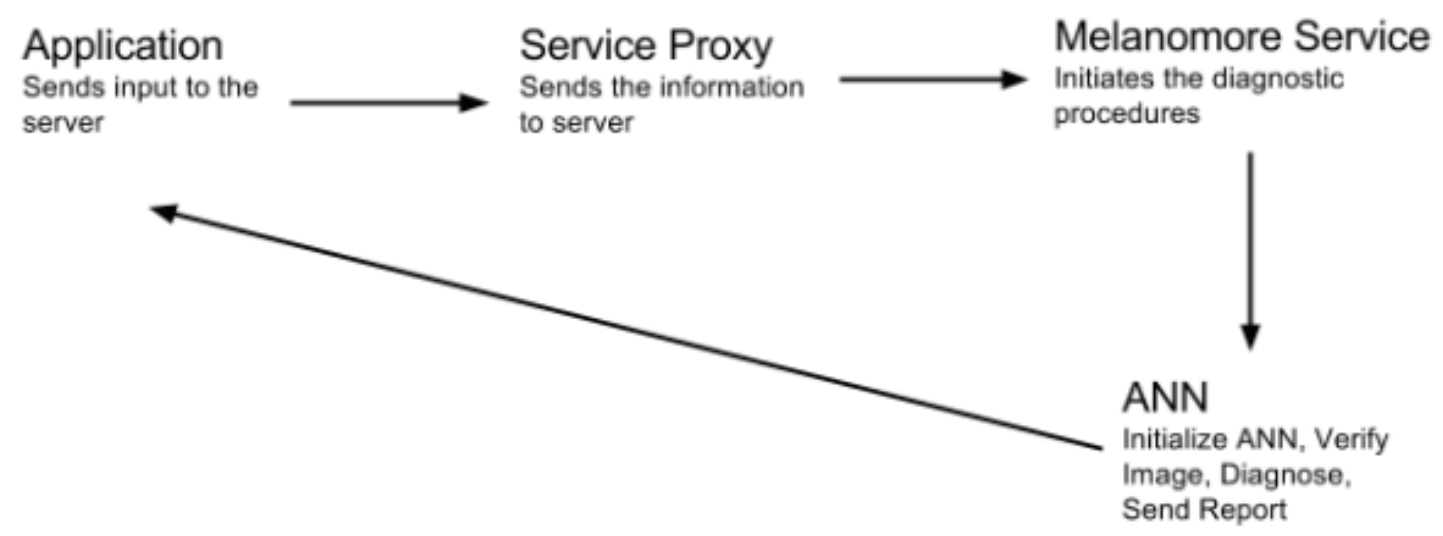

Figure 4: The flowchart of the processes of the application.

### 4.5 Results

The ANN employed in the prototype of the application had a promising accuracy of 92% For training and validating purposed over 450 images were used (400 from $PH^2$ dataset and 56 images sourced from other publicly available datasets) Throughout the training period, from an initial accuracy of 85% at use of 250 images, it gradually increased in accuracy to 93% at the end of 450 images used. Also, the FP, FN, TP, TN and the inconclusive rates were low.

A breakup of 70 : 30 was used for training and testing of our algorithm. The data presented here is from the testing period of the algorithm.

As the training size and the increased, the inconclusive diagnosis rate reduced too.

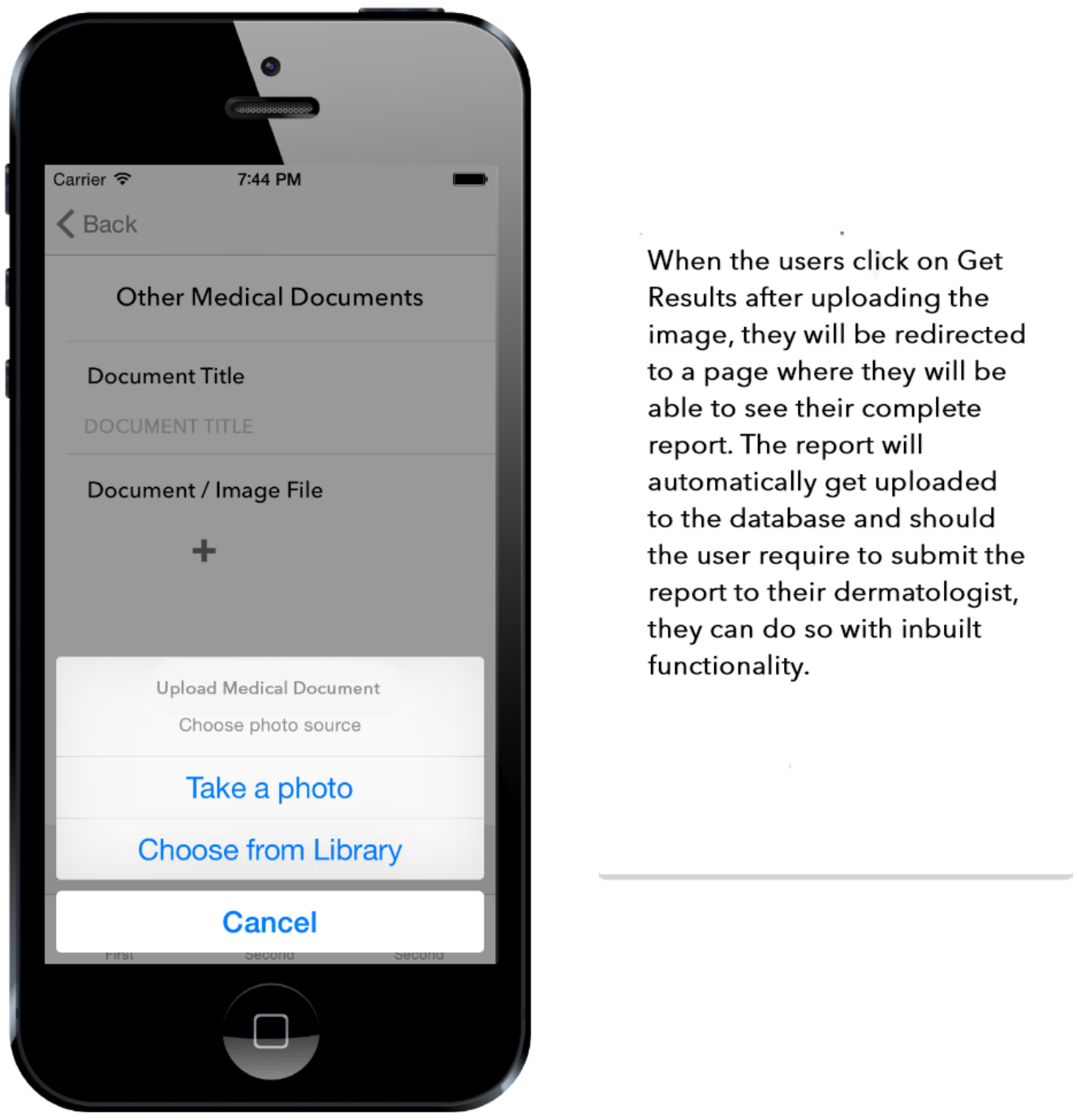

Figure 5: The mobile application for diagnosis of melanoma.

A T-test was taken to prove that the values calculated for asymmetry and border were actually different for both cases - malign as well as benign. The results of the T-test showed that both the groups were significantly different as both of them had extremely low probability values to be in the same group.

The results have been summarised below (Figure 8).

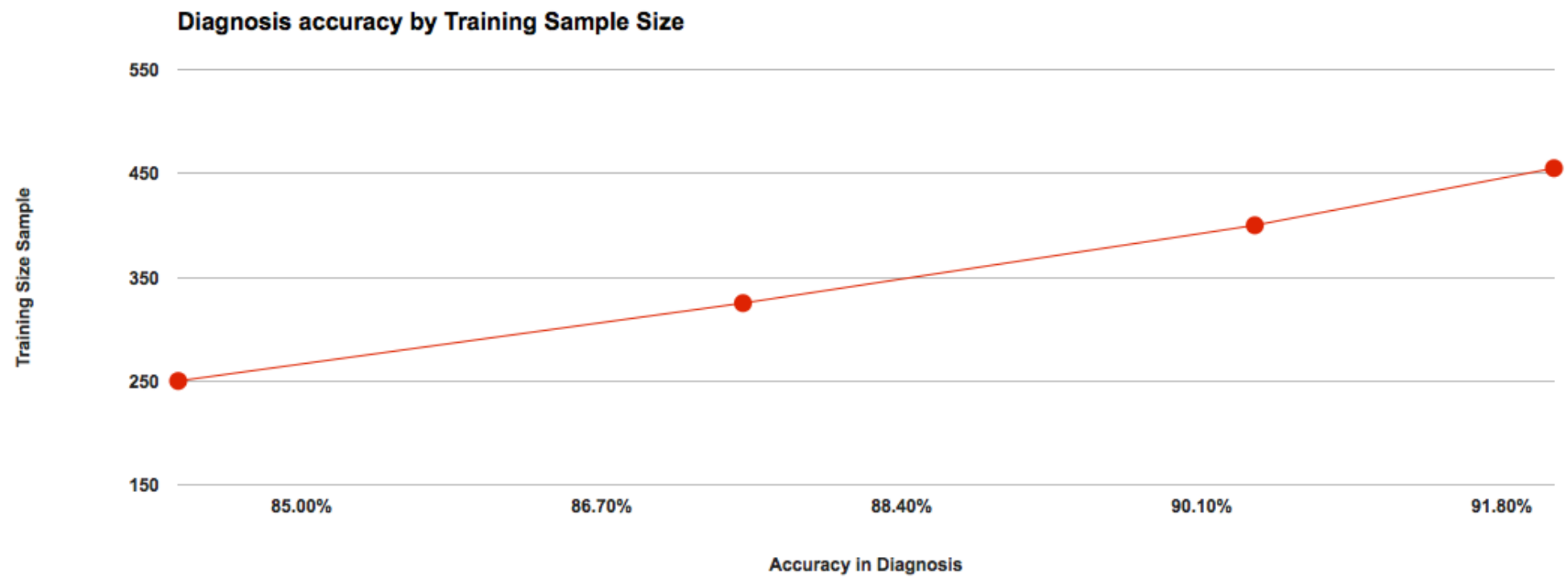

Figure 6: Comparison of training size and the accuracy of the algorithm.

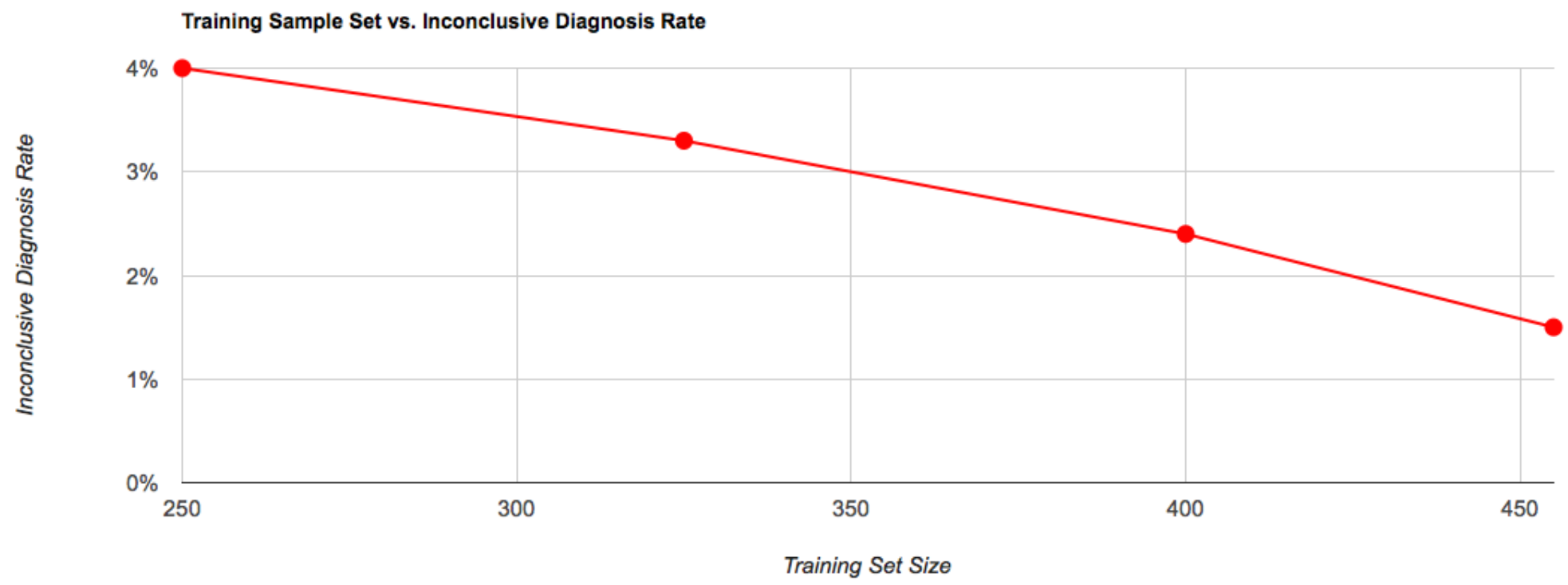

Figure 7: Comparison of training size and the inconclusive diagnosis rate of algorithm.

# 5 Breast Cancer

Artificial Neural Network and Gradient Descent were used in this part of the study.

## 5.1 Feature Selection

The features were selected from the Wisconsin Breast Cancer Database.

| **Feature** | **Details** |
|---|---|
| Clump Thickness | Assesses if cells are multi or mono-layered |
| Uniformity of Cell Size | Consistency in size of cells |
| Uniformity in Cell Shape | Estimation of equality of cell shapes and identification of marginal variances |
| Marginal Adhesion | Quantify the number of cells on the outside of epithelial that stick together |
| Single Epithelial Cell Size | Determines if epithelial cells are significantly enlarged |
| Bare Nuclei | Quantify the proportion of cells that are not surrounded by cytoplasm to those that are |
| Bland Chromatin | Rates uniformity of texture of nucleus |
| Normal Nucleoli | Determines whether nucleoli are small or large, barely visible or more visible |
| Mitoses | The level of mitotic acitivity |

### 5.2 Gradient Descent Development

A normalized gradient descent was developed in Python with custom libraries. Optimal results appeared at (number of iterations) $T = 5000$ and $\sigma = 2 \cdot 10^{-9}$.

### 5.3 Artificial Neural Network Development

A custom made ANN consisting of 7 layers was coded in Octave to process the dataset and achieve optimal results [25].

### 5.4 Results

Both the algorithms that were designed for diagnosing breast cancer delivered a high level of accuracy. The ANN rose in accuracy from 87% to around 95% on increase in training size. The gradient descent delivered an accuracy of 91% after 5000 iterations. The false positive and false negative rate in this case, was low too (Figure 9, Figure 10).

A breakup of 55 : 45 was used for training and testing of our algorithm. The data presented here is from the testing period of the algorithm.

| Overall Outcome | Overall Results | | | | |
|---|---|---|---|---|---|
| | Positive | | Negative | | Sensitivity 93.23% |
| | TP : 234 | FP : 15 | TN : 157 | FN : 17 | Specificity 91.28% |
| | Inconclusive : 6 | | | | Matthews Correlation Coefficient |
| | Positive Prediction Rate 93.98% | | Negative Prediction Rate 90.23% | | 0.844 |

Figure 8: Summary of the results from the Melanoma classifier.

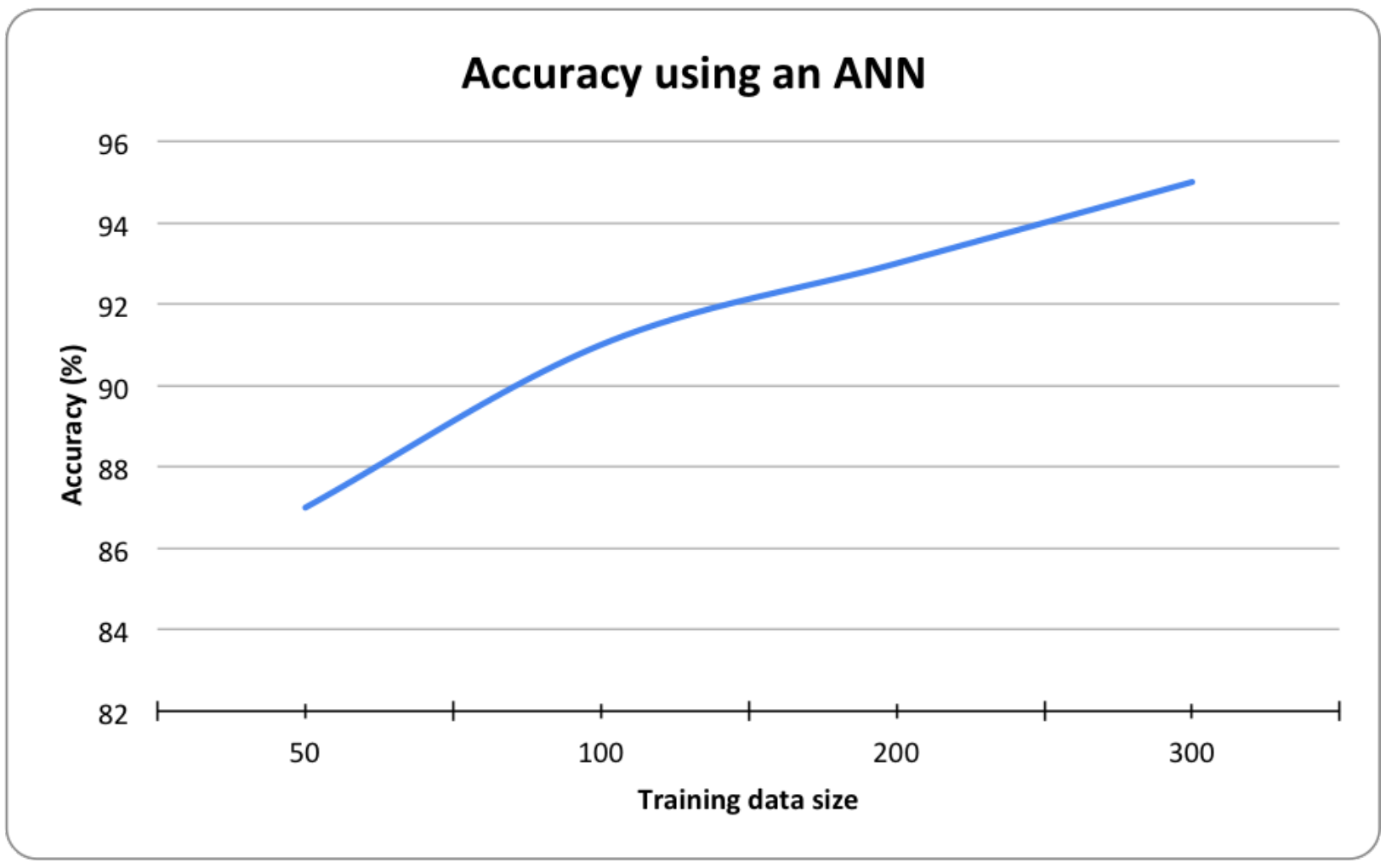

Figure 9: Comparison of the training data size and the accuracy of the algorithm.

The performance of the algorithms have been summarised (Figure 11).

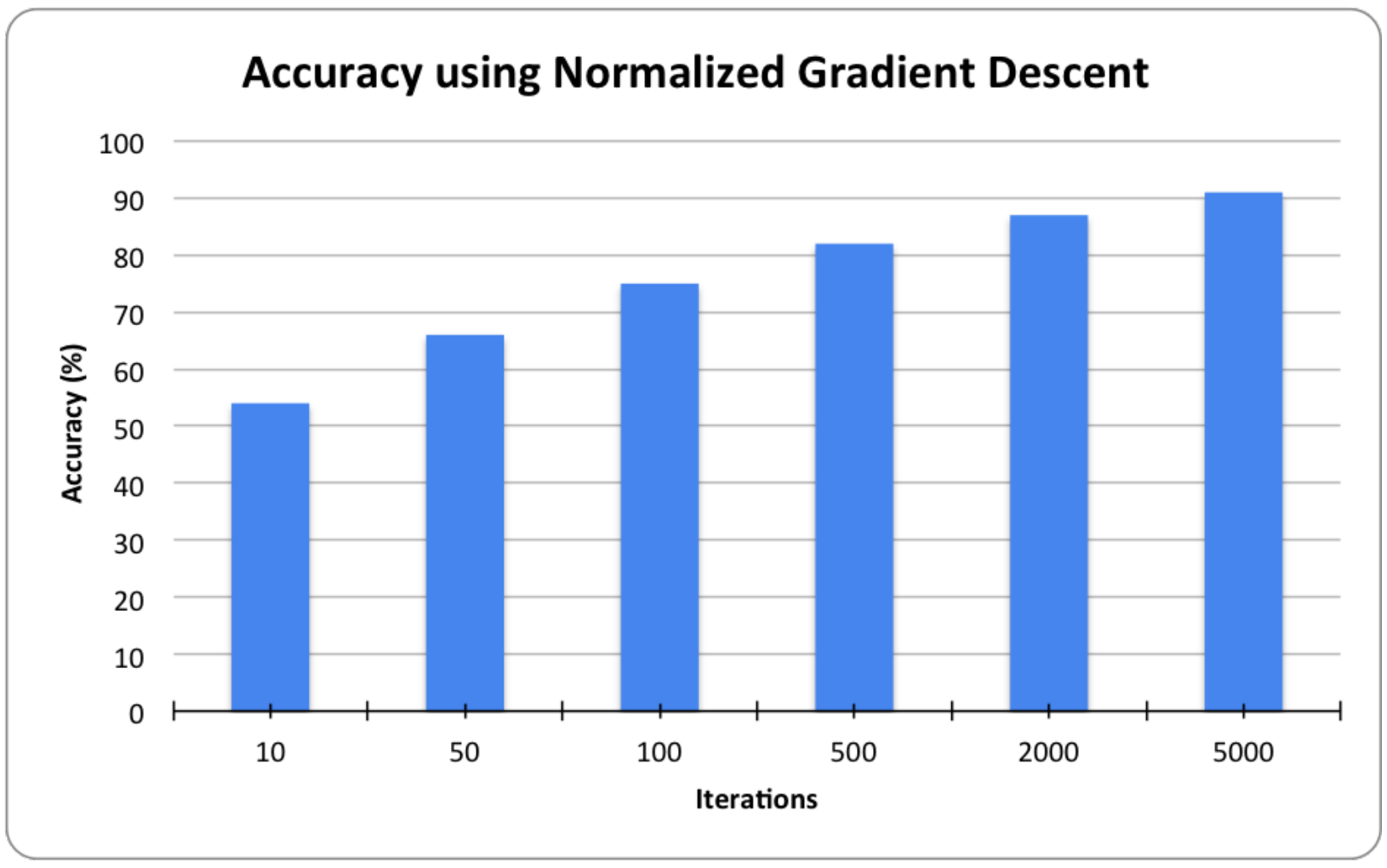

Figure 10: Comparison of the number of iterations and the accuracy of the algorithm.

| Overall Outcome | Overall Results | | | | |
|---|---|---|---|---|---|
| | Positive | | Negative | | Sensitivity 91.0% |
| | TP : 91 | FP : 5 | TN : 155 | FN : 9 | Specificity 96.8% |
| | Inconclusive : 0 | | | | Matthews Correlation Coefficient 0.886 |
| | Positive Prediction Rate 94.8% | | Negative Prediction Rate 94.4% | | |

Figure 11: Summary of the results from the Breast Cancer Classifier.

# 6 Conclusion

The algorithms achieved a high accuracy: The ANN used for detecting melanoma achieved an accuracy of 93%. The ANN and the Gradient Descent used for detecting breast cancer achieved an accuracy of 95% and 91% and the SVM used for detecting lung cancer achieved an accuracy of 94%. These algorithms also succeeded in achieving a very low false positive and false negative rate, indicating that our experiments were a success. All the algorithms used in this study were self developed. The ever learning nature of the algorithms make it possible for them to achieve near perfect accuracy on an increase in training size - something that an proposed online network facilitate.

Image processing and image analysis was used to obtain data directly from the CT scans and the skin images in the cases of lung cancer and melanoma. We decided to use the WBC dataset because it has been historically proven to be reliable. Images of the melanoma skin lesions as well as CT scans were taken from online databases (primarily $PH^2$ dataset and NCBI cancer imaging archive). We feel that the image processing algorithms gave us an advantage by allowing us to directly obtain data from the images. The success of our experiments also results due to our selection of features and methods for their calculations which were based on our exhaustive research of past literature.

Although previous work has succeeded in building computer aided systems for diagnosis (CAD) of diseases, CAD of cancer has only recently gained traction. The results achieved in this study are comparable to the seminal works in this field. In addition to the well-performing algorithms, image processing and analysis algorithms and feature selections, the construction of a naive mobile application in the case of melanoma diagnosis is something that people will greatly benefit from. The mobile application allows people directly access and take diagnostic tests on their mobile phones, hence eliminating costs associated with logistics and consultation. This is a boon in a world where consultation fees are sky-rocketing. In addition to this, a mobile application makes it feasible to carry out reliable tests in resources poor areas where taking exhaustive tests is not at all possible.

On observation, the algorithms used in this study also outperform several commercial software such as SkinSeg. With virtually no commercial software available for lung cancer and breast cancer diagnosis, we further plan to develop novel software that can be readily implemented in clinics all around the world. This will facilitate a global collaborative organisation that works on diagnosis research and also uses our software to actually diagnose diseases.

In the future, we plan to develop diagnostic algorithms for other diseases. We are also working on the online diagnosis software for collaborative research.

## 7 Acknowledgements

We would like to thank our family and friends for their tremendous support throughout the course of this project.

Mohnish would like to thank Giovanni Parmigiani, Chair and Professor, Department of Biostatistics and Computational Biology, Dana-Farber Cancer Institute for his invaluable advice during the early stages of this project.

Tanay would like to thank Philip Klein for providing the inspiration for this project through "Coding the Matrix: Linear Algebra through Computer Science Applications" and Andrew Ng for his instructional videos on "Machine Learning" on Coursera.

Finally, we would also like to thank the University of Porto, NCBI for letting us use their database.